\documentclass[journal]{IEEEtran}

\ifCLASSINFOpdf
\else
   \usepackage[dvips]{graphicx}
\fi
\usepackage{url}
\usepackage{graphicx}
\usepackage{amsmath}
\usepackage{amsfonts} 
\usepackage{multirow}
\usepackage{booktabs}
\usepackage{diagbox} 
\usepackage{flushend}
\usepackage{xcolor}
\begin{document}

\title{Vision-Language Consistency Guided Multi-modal Prompt Learning for Blind AI Generated Image Quality Assessment}

\author{Jun Fu, Wei Zhou, Qiuping Jiang, Hantao Liu, Guangtao Zhai
\thanks{This work was supported in part by Natural Science Foundation of China under Grant 62371305, 62001302. (Corresponding author: Wei Zhou.)}
\thanks{J. Fu is with the CAS Key Laboratory of Technology in Geo-Spatial Information Processing and Application System, University of Science and Technology of China, Hefei 230027, China (e-mail: fujun@mail.ustc.edu.cn).}
\thanks{W. Zhou and H. Liu are with the School of Computer Science and Informatics, Cardiff University, Cardiff CF24 4AG, United Kingdom (email: zhouw26@cardiff.ac.uk; liuh35@cardiff.ac.uk).}
\thanks{Q. Jiang is with the School of Information Science and Engineering, Ningbo University, Ningbo 315211, China (e-mail: jiangqiuping@nbu.edu.cn).}
\thanks{G. Zhai is with the Institute of Image Communication and Network Engineering, Shanghai Jiao Tong University, Shanghai 200240,
China (e-mail: zhaiguangtao@sjtu.edu.cn)}
}

\markboth{IEEE Signal Processing Letters}
{Shell \MakeLowercase{\textit{et al.}}: Bare Demo of IEEEtran.cls for IEEE Journals}
\maketitle

\begin{abstract}
Recently, textual prompt tuning has shown inspirational performance in adapting Contrastive Language-Image Pre-training (CLIP) models to natural image quality assessment. However, such uni-modal prompt learning method only tunes the language branch of CLIP models. This is not enough for adapting CLIP models to AI generated image quality assessment (AGIQA) since AGIs visually differ from natural images. In addition, the consistency between AGIs and user input text prompts, which correlates with the perceptual quality of AGIs, is not investigated to guide AGIQA. In this letter, we propose vision-language consistency guided multi-modal prompt learning for blind AGIQA, dubbed CLIP-AGIQA. Specifically, we introduce learnable textual and visual prompts in language and vision branches of CLIP models, respectively. Moreover, we design a text-to-image alignment quality prediction task, whose learned vision-language consistency knowledge is used to guide the optimization of the above multi-modal prompts. Experimental results on two public AGIQA datasets demonstrate that the proposed method outperforms state-of-the-art quality assessment models. The source code is available at https://github.com/JunFu1995/CLIP-AGIQA.

\end{abstract}

\begin{IEEEkeywords}
Multi-modal prompt learning, Vision-language consistency, AGIQA
\end{IEEEkeywords}

\IEEEpeerreviewmaketitle

\section{Introduction}
\IEEEPARstart{W}{ith} the rapid development of deep generation technology, we have entered the era of artificial intelligence (AI) generated content, where users can obtain images they want by feeding multiple text prompts into deep generative models. However, the quality of AI generated images (AGIs) is highly varied~\cite{zhang2023text}. Therefore, it is necessary to develop an objective image quality assessment (OIQA) metric to automatically filter out unqualified AGIs.

{In general, OIQA metrics encompass full-reference (FR) metrics, reduced-reference (RR) metrics, and blind metrics. FR metrics and RR metrics require referencing the original image, whereas blind metrics are reference-free. In real-world scenarios, the original AGI corresponding to user input text prompts is absent. Therefore, it is essential to develop blind IQA metrics in order to evaluate AGIs effectively.}

In the early stage, blind IQA metrics are designed based on handcrafted features, e.g., mean subtracted contrast normalized coefficients~\cite{brisque,niqe,ilniqe}, visual neuron matrix~\cite{chang2021blind}, and edge gradient features~\cite{feichtenhofer2013perceptual}. Since manually designing features is a time-consuming and error-prone process, researchers resort to convolutional neural networks~\cite{resnet} or transformers~\cite{liu2021swin}, and design  more sophisticated IQA models~\cite{cnniqa,hyperiqa,cheon2021perceptual}. Recently, Contrastive Language-Image Pre-training (CLIP) models are used to blindly assess the quality of natural images~\cite{clipiqa,msclip,zwx}, and shows inspirational zero-shot performance and potential to achieve competitive performance through textual prompt tuning~\cite{coop}. Motivated by the success of CLIP models in natural image quality assessment, we explore using CLIP models to assess the visual quality of AGIs in this letter. 

\begin{figure}[!tbp]
	\centering
	\includegraphics[width=\linewidth]{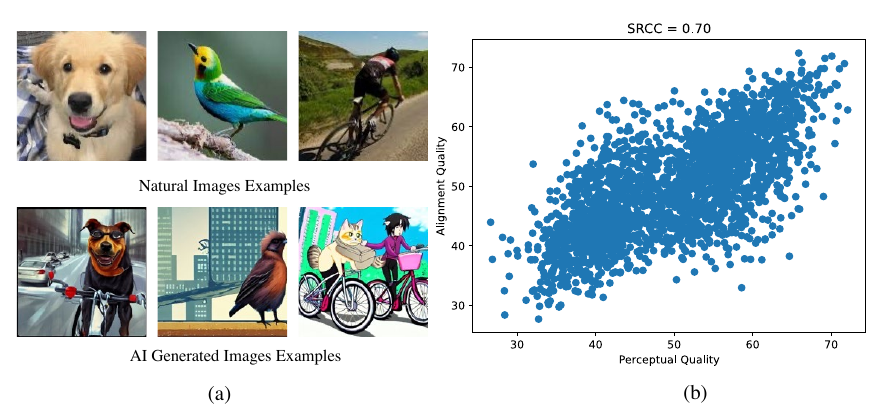} 
	\caption{(a) Comparison between natural images and AI generated images on the AGIQA-1K dataset~\cite{AGIQA1k}; (b) Spearman Rank Correlation Coefficient (SRCC) of the text-to-image alignment quality and perceptual quality on the AIGCIQA-2023 dataset~\cite{aigciqa2023}.}
\label{fig1}
\end{figure}

The marriage of CLIP models and AGIQA faces its unique challenges. First, besides textual prompt tuning, it needs to mitigate the domain gap between natural images and AGIs. As shown in Fig.~\ref{fig1}. (a), AGIs largely differ from natural images in terms of appearance and style. Second, it needs to explore using  vision-language consistency to guide AGIQA. As shown in Fig.~\ref{fig1}. (b), the alignment quality of the AGI and the user input text prompt is correlated with the perceived quality of the AGI. The reason for this phenomenon may be that users consider not only image fidelity when evaluating AGI, but also the consistency between the AGI and user input text prompts. Therefore, we believe that vision-language consistency is informative to the quality prediction of AGIs. 

{To tackle the aforementioned challenges, we propose a vision-language consistency guided multi-modal prompt learning approach.} Specifically, we add learnable prompts to both language and vision branches of CLIP models. In addition, we introduce a text-to-image alignment quality prediction task, whose learned vision-language consistency knowledge is used to guide the optimization of multi-modal prompts. In summary, {our contribution encompasses two distinct aspects}:
\begin{itemize}
	\item {As far as we know, we are the first one to explore CLIP models for blind AGIQA.}
	\item We study the use of text-to-image alignment information to assist the visual quality prediction of AGIs.
\end{itemize}

The remainder of this paper is organized as follows. Section II introduces the proposed method in detail. Section III provides experimental results and corresponding analysis. Finally, the paper is concluded in Section IV.

\begin{figure*}[!tbp]
	\centering
\includegraphics[width=0.85\linewidth]{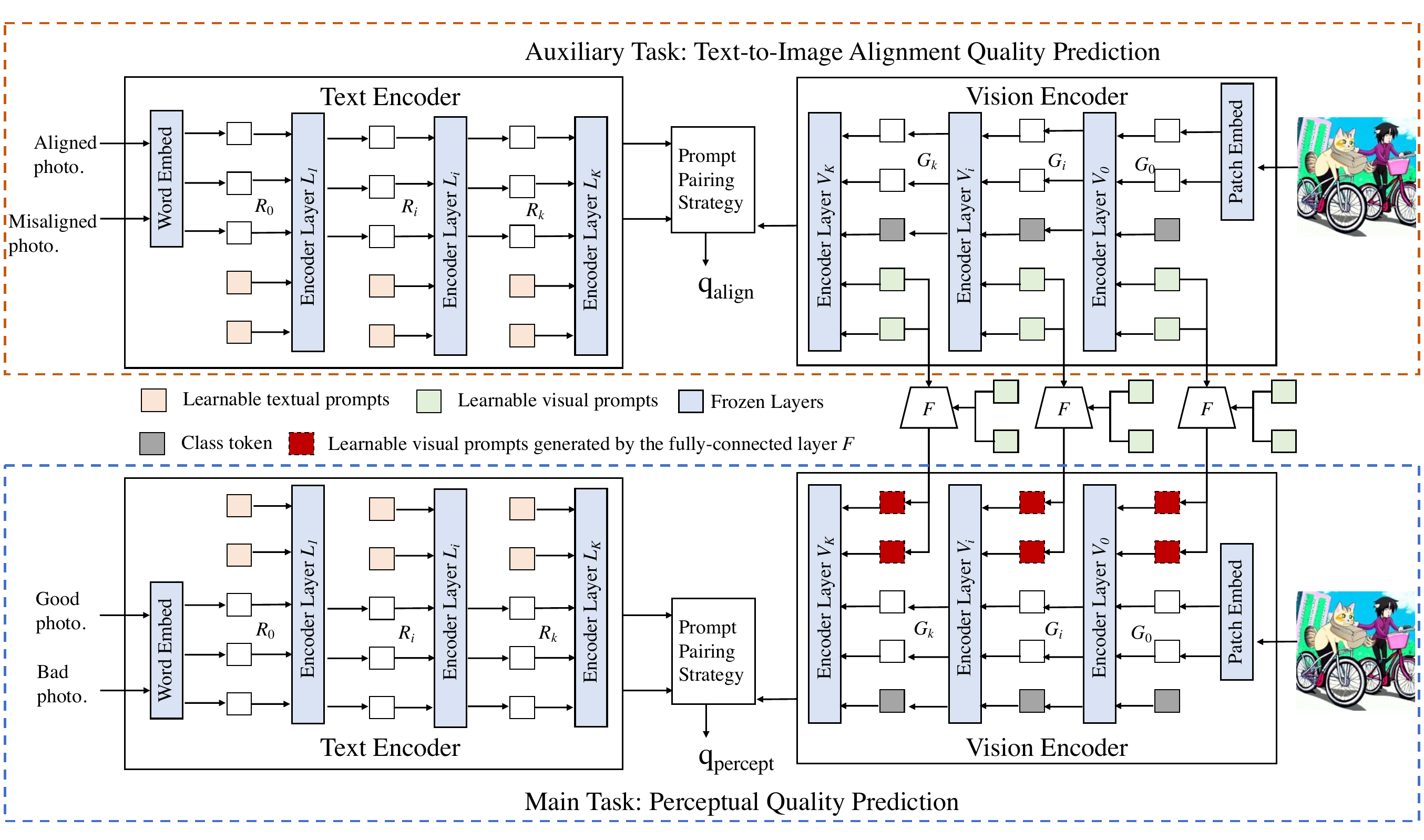} 
\caption{{Framework of our proposed method. It includes an auxiliary task and a main task. Both tasks are based on CLIP models and involve multi-modal prompt learning. Moreover, learnable visual prompts of the main task are conditioned on those of  the auxiliary task.}}
\label{fig:framework}
\end{figure*}

\section{Method}
Our approach, dubbed as CLIP-AGIQA, aims to exploit multi-modal prompt learning to fine-tune CLIP models. Unlike previous methods~\cite{maple,xing2023dual} that optimize multi-modal prompts with only the target task, we introduce an auxiliary task to guide the multi-modal prompt learning.  Fig.~\ref{fig:framework} shows the overall architecture of our proposed framework. As we can see, {our approach comprises a perceptual quality prediction task and a text-to-image alignment quality prediction task.} Both tasks adopt multi-modal prompt learning to finetune CLIP models. Moreover, there are interactions between the learnable prompts of two tasks. During fine-tuning, the CLIP model is frozen while the rest of the proposed framework is optimized. Below, we first recap the CLIP architecture and then detail the proposed framework.

\subsection{Recap of CLIP Models}
Referring to previous prompting methods~\cite{coop,CoCOOP,maple}, {here we adopt} transformer-based CLIP models. In the CLIP model,  {vision and text encoders} are used to generate image and text representations, respectively. The details are introduced below.

For the vision encoder $\mathcal{V}$, the input image $I$ is divided into $M$ fixed-size patches, and each patch is projected into $d_v$-dimensional latent space. The resulting patch embeddings $G_0 \in \mathbb{R}^{M \times d_v}$ and a learnable class token $\text{c}_0 \in \mathbb{R}^{d_{v}} $ are fed into the transformer block $\mathcal{V}_{0}$, which later is repeated $k-1$ times. The whole process can be formulated as,
\begin{equation}
[{\text{c}}_i, G_i] = \mathcal{V}_{i}([\text{c}_{i-1}, G_{i-1}])  ~~~~ i= 1, \cdots, k.
\end{equation} 
The image representation $x \in \mathbb{R}^{d_{vl}}$is obtained by projecting the class token $\text{c}_{k}$  into $d_{vl}$-dimensional latent space.

For the text encoder $\mathcal{L}$, the input text description is tokenized into words, and each word is projected into $d_l$-dimensional latent space. The resulting word representations ${R}_0 =[r_{0}^{1}, \cdots, r_{0}^{N}] \in \mathbb{R}^{N \times d_{l}}$ are sequentially processed by $k$ transformer layers, formulated as,
\begin{equation}
[{R}_i] = \mathcal{L}_{i}({R}_{i-1}) \ \ \ ~~~~ i= 1,  \cdots, k.
\end{equation}
The text representation $z \in \mathbb{R}^{d_{vl}}$is obtained by projecting the last token $r_{k}^{N}$ into the same space as the image representation.

\subsection{Text-to-Image Alignment Quality Prediction}
In the AGIQA dataset, the human-annotated text-to-image alignment scores, which reflect the consistency between AGIs and corresponding user input text prompts, are typically available. Since alignment scores are correlated with the perceptual quality of AGIs, we aim to learn the vision-language consistency knowledge to help AGI quality assessment. 

A straightforward approach is to add learnable prompts into the vision encoder of the CLIP model and optimize them towards making the similariy between the AGI and user input text prompts close to the alignment score.  However, the user input text prompts sometimes only contain several keywords (e.g., the AGIQA-1k dataset~\cite{AGIQA1k}), which are not informative. In addition, the user input text prompt is absent in some AGIQA datasets, e.g., the AIGCIQA-2023 dataset~\cite{aigciqa2023}. Therefore, we explore a blind setting, where we predict the alignment score without the user input text prompts. 

Specifically, we use a prompt pairing strategy to estimate the alignment score  of the AGI. Let us denote $t^{align}_1$ and $t^{align}_2$ as a pair of antonym prompts, i.e., ``\texttt{Aligned photo.}'' and ``\texttt{Misaligned photo.}''. We first compute the cosine similarity  between  manually designed antonym prompts and the AGI as follows,
\begin{equation}
s^{align}_i = \frac{\mathcal{V}(I) \odot \mathcal{L}(t^{align}_i)}{\lVert\mathcal{V}(I)\rVert \cdot \lVert\mathcal{L}(t^{align}_i)\rVert}, i\in \{1, 2\},
\end{equation}
where $\lVert \cdot \rVert$ denotes $l_2$ norm and $\odot$ represents the vector dot-product. Then, we estimate the alignment score as follows, 
\begin{equation}
q_{align} = \frac{e^{s^{align}_1}}{e^{s^{align}_1}+e^{s^{align}_2}}.
\end{equation}
Since hand-crafted antonym prompts are often not optimal, we introduce $b$ learnable prompts $P^{align}_{i-1} \in \mathbb{R}^{b\times d_l}$ into each transformer layer of the text encoder, formulated as,
\begin{align}
[\ \underline{\hspace{0.3cm}}\ , R_{i}] = \mathcal{L}_{i}([P^{align}_{i-1}, R_{i-1}]) ~~~~ i=1,  \cdots, k.
\end{align}
In addition, since CLIP models, pretrained on natural images, are limited to capture distinguishable image representations for AGIs,  we also introduce $b$ learnable prompts $Q^{align}_{i-1} \in \mathbb{R}^{b\times d_l}$ into each transformer layer of the vision encoder, formulated as, 
\begin{equation}
[c_i, G_i, \ \underline{\hspace{0.3cm}} \ ] = \mathcal{V}_{i}([c_{i-1}, G_{i-1}, {Q}^{align}_{i-1}])  \ ~ i=1,  \cdots, k. \\
\end{equation}

\subsection{Perceptual Quality Prediction}
Like the text-to-image alignment quality prediction task, we also use the prompt pairing strategy to estimate the perceptual quality of the AGI. Let us denote $t_1^{percept}$ and $t_2^{percept}$ as ``\texttt{Good photo.}'' and ``\texttt{Bad photo.}'', respectively.  The predicted perceptual quality is computed as follows,
\begin{equation}
q_{percept} = \frac{e^{s_1^{percept}}}{e^{s_1^{percept}}+e^{s_2^{percept}}}. 
\end{equation}
 In addition, we also adopt multi-modal prompt learning to fine-tune the CLIP model. Specifically, the formulation of the text encoder is defined as,
\begin{equation}
[\ \underline{\hspace{0.3cm}}, \ R_{i}] = \mathcal{L}_{i}([P^{percept}_{i-1}, R_{i-1}]) ~~i=1,  \cdots, k,
\end{equation}
where $P^{percept}_{i-1}  \in \mathbb{R}^{b\times d_l} $ is learnable textual prompts. 
The vision encoder can expressed as, 
\begin{equation}
\small
\begin{split}
[c_i, G_i, \ \underline{\hspace{0.3cm}} \ ] &= \mathcal{V}_{i}([c_{i-1}, G_{i-1}, \tilde{Q}^{percept}_{i-1}])  \ ~ i=1, \cdots, k, \\
\tilde{Q}^{percept}_{i-1} &= \mathcal{F}_{i-1}([Q^{align}_{i-1}, Q^{percept}_{i-1}]) \ ~ i=1,  \cdots, k,
\end{split}
\label{eq:1}
\end{equation}
where $\mathcal{F}_{i-1}$ denotes the fully-connected layer, $Q^{percept}_{i-1}$ represents learnable visual prompts in the perceptual quality prediction task. As shown in Equation \ref{eq:1}, we explicitly condition $Q^{percept}_{i-1}$ on the learnable visual prompts $Q^{align}_{i-1}$ in the text-to-image alignment score prediction task. The motivation behind this is that  $Q^{align}_{i-1}$ contains the vision-language consistency knowledge which is informative to the perceptual quality prediction. Notably, we empirically find that adding such conditions to textual learnable prompts brings limited gains.

\subsection{Loss Function}
The loss function for the alignment score prediction is defined as, 
\begin{equation}
L_{align} = \frac{1}{N} \sum_{i=1}^{N}\lVert q^i_{align} - g_{align}^i\rVert^2_2,
\end{equation}
where $N$ is the batch size and $g_{align}^i$ is the ground-truth alignment score of $i$-th AGI. The loss function for the perceptual quality prediction is defined as, 
\begin{equation}
L_{percept} = \frac{1}{N} \sum_{i=1}^{N}\lVert q^i_{percept} - g_{percept}^i\rVert^2_2,
\end{equation}
where $g_{percept}^i$ is the ground-truth perceptual quality of $i$-th AGI. The final loss function is defined as, 
\begin{equation}
L =L_{percept} + \lambda L_{align},
\end{equation}
where $\lambda$ is a hyperparameter.

\section{Experiments}
\subsection{Database and Evaluation Criteria}
We conduct extensive experiments on two public AGIQA datasets, i.e., AGIQA-3K~\cite{agiqa3k} and AIGCIQA-2023~\cite{aigciqa2023}. The AGIQA-3K database contains 2982 AGIs which are generated by Glide~\cite{glide}, Stable Diffusion~\cite{sd}, Stable Diffusion
XL~\cite{rombach2022text}, Midjourney~\cite{md}, AttnGAN\cite{attngan}, and DALLE2~\cite{dalle2}. The AIGCIQA-2023 dataset {uses} Glide~\cite{glide}, Lafite~\cite{zhou2022towards}, DALLE2~\cite{dalle2}, Stable Diffusion~\cite{sd}, Unidiffuser~\cite{bao2023one}, and Controlnet~\cite{controlnet} to generate 2400 AGIs. In both datasets, each AGI is accompanied with a perceptual quality and alignment score, which are annotated by subjects. Notably, the user input text prompts are not available in the AIGCIQA-2023 dataset.

{We use Spearman Rank Correlation Coefficient (SRCC), Pearson Linear Correlation Coefficient (PLCC), and Kendall’s Rank Correlation Coefficient (KRCC) to compare IQA metrics.} Good IQA methods generally achieve high scores in all three evaluation metrics. Since the AIGCIQA dataset is limited in scale, we evaluate each IQA model 10 times, and report the average performance.

\subsection{Implement Details}
We use a ViT-B/32 based CLIP model, where the length of the learnable multi-modal prompt is set to 8. The hyperparameter $\lambda$ is empirically set to 0.1. {The dataset is partitioned into training and testing sets at an 8:2 ratio, ensuring that images with the same user prompts are grouped together. During the training process, 64 patches with a size of 224 $\times$ 224 are fed into the CLIP model at each iteration. We employ Adam algorithm~\cite{adam} to optimize the learnable parameters of the model. The learning rate and training epoch are configured as 1e-4 and 50, respectively. During the testing,  we calculate the quality score of the input AGI using a patch-based evaluation fashion~\cite{fu2023scale,zhou2020blind,zhou2019dual}.}  {We implement our method based on PyTorch~\cite{paszke2019pytorch}, and run all experiments on a NVIDIA RTX 4090 GPU platform with an  Intel Core i9-13900KF CPU.} 

\subsection{Performance Comparisons}
{To validate the efficacy of the proposed approach, we conduct a comparative analysis with} three hand-crafted feature based methods~\cite{brisque,niqe,ilniqe}, three deep-learning based approaches~\cite{resnet,cnniqa,hyperiqa}, and two CLIP based metrics~\cite{clipiqa}. {The results are reported in Table~\ref{tab:1}. Based on the data provided in Table~\ref{tab:1}, we can draw the following conclusions.} First, handcrafted feature based methods achieve poor performance on both AGIQA datasets. This is because AGIs largely differ from natural images for which handcrafted features are designed. {Second, the deep learning-based methods achieve relatively higher accuracy.} This verifies the superiority of learned features over handcrafted ones. Third, CLIPIQA shows impressive zero-shot performance, and CLIPIQA$^{+}$ further improves the performance through textual prompt learning. This shows the promising potential of exploring CLIP models for AGIQA. Lastly, the proposed method called CLIP-AGIQA shows a clear advantage over CLIPIQA$^{+}$ on both datasets. This confirms the effectiveness of the proposed method.

\begin{table}[t]
\centering
\caption{Performance comparisons of objective quality metrics on AGIQA-3K and AIGCIQA-2023 databases. }
\scalebox{0.7}{
\begin{tabular}{|l|l|lll|lll|}
\hline
\multirow{2}{*}{Type} & \multirow{2}{*}{Method} &  \multicolumn{3}{c|}{AGIQA-3k Database} &  \multicolumn{3}{c|}{AIGCIQA-2023 Database} \\  \cline{3-8}
& & SRCC & PLCC & KRCC & SRCC & PLCC & KRCC  \\ \hline
\multirow{3}{*}{\begin{tabular}[c]{@{}l@{}}Handcrafted\\ feature\\ based\end{tabular}} &    BRISQUE~\cite{brisque}  & 0.4932&0.5399&0.3348   & 0.6309&0.5977&0.4348 \\

& NIQE~\cite{niqe}   &  0.5151&0.5241&0.3499   & 0.4870&0.4576&0.3270  \\
&  ILNIQE~\cite{ilniqe}  &  0.5935&0.6240&0.4183   &  0.5576&0.4933&0.3762  \\ \hline 

\multirow{3}{*}{\begin{tabular}[c]{@{}l@{}}Deep\\ learning-\\ based\end{tabular}} &  CNNIQA~\cite{cnniqa}  & 0.7437 &  0.8332&  0.5516 & 0.6974&0.7011&0.4873 \\
&  ResNet50~\cite{resnet}  &0.8445&0.9033&0.6631  & 0.8113&0.8416&0.5956  \\
&  HyperIQA~\cite{hyperiqa}  & 0.8433&0.9013&0.6612   &  0.8174&0.8459&0.6032  \\\hline 
\multirow{3}{*}{\begin{tabular}[c]{@{}l@{}}CLIP\\ based \end{tabular}} &  CLIPIQA~\cite{clipiqa}  &  0.6846&0.6987&0.4915  &  0.4171&0.3970&0.2823  \\
&  CLIPIQA$^{+}$~\cite{clipiqa}  &0.8428&0.8879&0.6556   &  0.8072&0.8280&0.5905  \\
&  CLIP-AGIQA  & \textbf{0.8747}&\textbf{0.9190}&\textbf{0.6976}   & \textbf{0.8324}&\textbf{0.8604}&\textbf{0.6220} \\\hline 
\end{tabular}
}
\label{tab:1}
\end{table}
\begin{table}[htbp] 
	\centering
	\caption{ Ablation Study on each component of the proposed method. The training time and testing time are calculated on images of spatial size 224 $\times$ 224. }
	\scalebox{0.75}{
		\begin{tabular}{|l|c|c|c|c|}
			\hline 
			\diagbox{Components}{Method} & CLIPIQA & $A_1$ & $A_2$ & CLIP-AGIQA\\	\hline 
			Handcrafted Text Prompts &  \checkmark & & &\\
			Textual Prompt Learning &  & \checkmark& \checkmark& \checkmark\\ 
			Visual Prompt Learning & & & \checkmark & \checkmark\\ 		
			Vision-Language Consistency  &  & &  & \checkmark \\ 		
			\hline 
			SRCC & 0.6846 & 0.8473&0.8696 & \textbf{0.8747}  \\ 		
			PLCC & 0.6987 &0.8929&0.9186 & \textbf{0.9190}  \\ 		
			KRCC & 0.4915 & 0.6611&0.6919 &\textbf{0.6976}  \\ 		\hline
			Training time per epoch (s)& \textbf{0} & 5.560 & 6.823& 8.062\\
			Testing time per epoch (s)& \textbf{5.531} & 5.567 & 5.617& 5.772\\\hline 
	\end{tabular}}
	\label{tab:2}
\end{table}
\subsection{Ablation Study}

{We first evaluate the efficacy of each component in the proposed method. The findings are presented in Table ~\ref{tab:2}, from which we can infer the following conclusions.} First, the variant method $A_1$, which only uses textual prompt learning, achieves better performance than CLIPIQA which uses handcrafted text prompts. This confirms the necessity of using textual prompt learning. Second, the variant method $A_2$, which uses textual and visual prompt learning, is superior to $A_1$. This confirms the advantage of multi-modal prompt learning over textual prompt learning. Third, the proposed method slightly outperforms $A_2$. This shows that the vision-language consistency knowledge is informative to AGIQA. {Fourth, while the proposed method has much higher training cost than CLIPIQA which does not require training, its inference cost is comparable to CLIPIQA's. This is mainly because the auxiliary task can be discarded in the testing phase.}

\begin{table}[htbp] 
	\centering
	\caption{ The impact of  vision-Language consistency on the performance.}
	\scalebox{0.8}{
		\begin{tabular}{|l|c|c|c|}
			\hline 
			\diagbox{Method}{Metrics} & SRCC  & PLCC & KRCC \\	\hline  
			$B_1$  & 0.8741 & 0.9182 & 0.6965 \\
			CLIP-AGIQA & \textbf{0.8747} & \textbf{{0.9190}}  & \textbf{{0.6976}}\\\hline
	\end{tabular}}
	\label{tab:3}
\end{table} 
{Subsequently, we conduct an investigation into the vision-language consistency.} The results are shown in Table~\ref{tab:3}. $B_1$ adopts the same framework as the proposed method, while learning the vision-language consistency knowledge with user input text prompts. More specifically,  in the text-to-image quality prediction task, it feeds user input text descriptions into the text encoder without learnable textual prompts. {According to Table~\ref{tab:3}, we find that the proposed method exhibits a slight superiority over $B_1$.} {The possible reason for this phenomenon is as follows. For the text-to-image alignment quality prediction, the blind setting (i.e., without user input text prompts) is usually more challenging than the non-blind setting (i.e., with user input text prompts), which helps us learn non-trivial vision-language consistency knowledge.}

\begin{table}[htbp] 
	\centering
	\caption{ Comparison with competitive prompt learning methods.}
		\scalebox{0.7}{
	\begin{tabular}{|l|c|c|c|}
		\hline 
		\diagbox{Metrics}{Method} & CoCoOP~\cite{CoCOOP} & MaPLe~\cite{maple} & CLIP-AGIQA \\	\hline  
		SRCC &0.8582&  0.8713  & \textbf{0.8747} \\ 		
		PLCC & 0.9079& 0.9176 & \textbf{0.9190} \\ 		
		KRCC & 0.6759& 0.6939 & \textbf{0.6976} \\ 		\hline 
	\end{tabular}}
	\label{tab:4}
\end{table}  
{Finally, we compare the proposed method with two competitive prompt learning methods, i.e., CoCoOP~\cite{CoCOOP} and MaPLe~\cite{maple}. For fair comparison, these two methods share the same training and testing settings as our proposed method.} The results are presented in Table~\ref{tab:4}. As shown, the proposed method exceeds CoCoOP by a clear margin. Moreover, the proposed method slightly outperforms the multi-modal prompt learning approach known as MaPLe. {This can be attributed to the learned vision-language consistency knowledge.}

\section{Conclusion}
In this letter,  we propose vision-language consistency guided multi-modal prompt learning to adapt CLIP models to blindly assess the visual quality of AI generated images. { Experiments evince that our approach achieves more accurate predictions than existing IQA metrics, and each technical component in our method plays a crucial role.} {However, since the auxiliary task is designed as text-to-image alignment quality prediction, our method cannot be applied to the scenario where alignment quality scores are unavailable. Therefore, we will explore better auxiliary tasks in the future.}

\bibliographystyle{IEEEtran}
\bibliography{references}

\begin{thebibliography}{10}
\providecommand{\url}[1]{#1}
\csname url@samestyle\endcsname
\providecommand{\newblock}{\relax}
\providecommand{\bibinfo}[2]{#2}
\providecommand{\BIBentrySTDinterwordspacing}{\spaceskip=0pt\relax}
\providecommand{\BIBentryALTinterwordstretchfactor}{4}
\providecommand{\BIBentryALTinterwordspacing}{\spaceskip=\fontdimen2\font plus
\BIBentryALTinterwordstretchfactor\fontdimen3\font minus
  \fontdimen4\font\relax}
\providecommand{\BIBforeignlanguage}[2]{{%
\expandafter\ifx\csname l@#1\endcsname\relax
\typeout{** WARNING: IEEEtran.bst: No hyphenation pattern has been}%
\typeout{** loaded for the language `#1'. Using the pattern for}%
\typeout{** the default language instead.}%
\else
\language=\csname l@#1\endcsname
\fi
#2}}
\providecommand{\BIBdecl}{\relax}
\BIBdecl

\bibitem{zhang2023text}
C.~Zhang, C.~Zhang, M.~Zhang, and I.~S. Kweon, ``Text-to-image diffusion model
  in generative ai: A survey,'' \emph{arXiv preprint arXiv:2303.07909}, 2023.

\bibitem{brisque}
A.~Mittal, A.~K. Moorthy, and A.~C. Bovik, ``No-reference image quality
  assessment in the spatial domain,'' \emph{IEEE Transactions on image
  processing}, vol.~21, no.~12, pp. 4695--4708, 2012.

\bibitem{niqe}
A.~Mittal, R.~Soundararajan, and A.~C. Bovik, ``Making a “completely blind”
  image quality analyzer,'' \emph{IEEE Signal processing letters}, vol.~20,
  no.~3, pp. 209--212, 2012.

\bibitem{ilniqe}
L.~Zhang, L.~Zhang, and A.~C. Bovik, ``A feature-enriched completely blind
  image quality evaluator,'' \emph{IEEE Transactions on Image Processing},
  vol.~24, no.~8, pp. 2579--2591, 2015.

\bibitem{chang2021blind}
H.-W. Chang, X.-D. Bi, and C.~Kai, ``Blind image quality assessment by visual
  neuron matrix,'' \emph{IEEE Signal Processing Letters}, vol.~28, pp.
  1803--1807, 2021.

\bibitem{feichtenhofer2013perceptual}
C.~Feichtenhofer, H.~Fassold, and P.~Schallauer, ``A perceptual image sharpness
  metric based on local edge gradient analysis,'' \emph{IEEE Signal Processing
  Letters}, vol.~20, no.~4, pp. 379--382, 2013.

\bibitem{resnet}
K.~He, X.~Zhang, S.~Ren, and J.~Sun, ``Deep residual learning for image
  recognition,'' in \emph{Proceedings of the IEEE conference on computer vision
  and pattern recognition}, 2016, pp. 770--778.

\bibitem{liu2021swin}
Z.~Liu, Y.~Lin, Y.~Cao, H.~Hu, Y.~Wei, Z.~Zhang, S.~Lin, and B.~Guo, ``Swin
  transformer: Hierarchical vision transformer using shifted windows,'' in
  \emph{Proceedings of the IEEE/CVF international conference on computer
  vision}, 2021, pp. 10\,012--10\,022.

\bibitem{cnniqa}
L.~Kang, P.~Ye, Y.~Li, and D.~Doermann, ``Convolutional neural networks for
  no-reference image quality assessment,'' in \emph{Proceedings of the IEEE
  conference on computer vision and pattern recognition}, 2014, pp. 1733--1740.

\bibitem{hyperiqa}
S.~Su, Q.~Yan, Y.~Zhu, C.~Zhang, X.~Ge, J.~Sun, and Y.~Zhang, ``Blindly assess
  image quality in the wild guided by a self-adaptive hyper network,'' in
  \emph{Proceedings of the IEEE/CVF Conference on Computer Vision and Pattern
  Recognition}, 2020, pp. 3667--3676.

\bibitem{cheon2021perceptual}
M.~Cheon, S.-J. Yoon, B.~Kang, and J.~Lee, ``Perceptual image quality
  assessment with transformers,'' in \emph{Proceedings of the IEEE/CVF
  Conference on Computer Vision and Pattern Recognition}, 2021, pp. 433--442.

\bibitem{clipiqa}
J.~Wang, K.~C. Chan, and C.~C. Loy, ``Exploring clip for assessing the look and
  feel of images,'' in \emph{Proceedings of the AAAI Conference on Artificial
  Intelligence}, vol.~37, no.~2, 2023, pp. 2555--2563.

\bibitem{msclip}
T.~Miyata, ``Interpretable image quality assessment via clip with multiple
  antonym-prompt pairs,'' \emph{arXiv preprint arXiv:2308.13094}, 2023.

\bibitem{zwx}
W.~Zhang, G.~Zhai, Y.~Wei, X.~Yang, and K.~Ma, ``Blind image quality assessment
  via vision-language correspondence: A multitask learning perspective,'' in
  \emph{Proceedings of the IEEE/CVF Conference on Computer Vision and Pattern
  Recognition}, 2023, pp. 14\,071--14\,081.

\bibitem{coop}
K.~Zhou, J.~Yang, C.~C. Loy, and Z.~Liu, ``Learning to prompt for
  vision-language models,'' \emph{International Journal of Computer Vision},
  vol. 130, no.~9, pp. 2337--2348, 2022.

\bibitem{AGIQA1k}
Z.~Zhang, C.~Li, W.~Sun, X.~Liu, X.~Min, and G.~Zhai, ``A perceptual quality
  assessment exploration for aigc images,'' \emph{arXiv preprint
  arXiv:2303.12618}, 2023.

\bibitem{aigciqa2023}
J.~Wang, H.~Duan, J.~Liu, S.~Chen, X.~Min, and G.~Zhai, ``Aigciqa2023: A
  large-scale image quality assessment database for ai generated images: from
  the perspectives of quality, authenticity and correspondence,'' \emph{arXiv
  preprint arXiv:2307.00211}, 2023.

\bibitem{maple}
M.~U. Khattak, H.~Rasheed, M.~Maaz, S.~Khan, and F.~S. Khan, ``Maple:
  Multi-modal prompt learning,'' in \emph{Proceedings of the IEEE/CVF
  Conference on Computer Vision and Pattern Recognition}, 2023, pp.
  19\,113--19\,122.

\bibitem{xing2023dual}
Y.~Xing, Q.~Wu, D.~Cheng, S.~Zhang, G.~Liang, P.~Wang, and Y.~Zhang, ``Dual
  modality prompt tuning for vision-language pre-trained model,'' \emph{IEEE
  Transactions on Multimedia}, 2023.

\bibitem{CoCOOP}
K.~Zhou, J.~Yang, C.~C. Loy, and Z.~Liu, ``Conditional prompt learning for
  vision-language models,'' in \emph{Proceedings of the IEEE/CVF Conference on
  Computer Vision and Pattern Recognition}, 2022, pp. 16\,816--16\,825.

\bibitem{agiqa3k}
C.~Li, Z.~Zhang, H.~Wu, W.~Sun, X.~Min, X.~Liu, G.~Zhai, and W.~Lin,
  ``Agiqa-3k: An open database for ai-generated image quality assessment,''
  \emph{arXiv preprint arXiv:2306.04717}, 2023.

\bibitem{glide}
A.~Nichol, P.~Dhariwal, A.~Ramesh, P.~Shyam, P.~Mishkin, B.~McGrew,
  I.~Sutskever, and M.~Chen, ``Glide: Towards photorealistic image generation
  and editing with text-guided diffusion models,'' \emph{arXiv preprint
  arXiv:2112.10741}, 2021.

\bibitem{sd}
R.~Rombach, A.~Blattmann, D.~Lorenz, P.~Esser, and B.~Ommer, ``High-resolution
  image synthesis with latent diffusion models,'' in \emph{Proceedings of the
  IEEE/CVF conference on computer vision and pattern recognition}, 2022, pp.
  10\,684--10\,695.

\bibitem{rombach2022text}
R.~Rombach, A.~Blattmann, and B.~Ommer, ``Text-guided synthesis of artistic
  images with retrieval-augmented diffusion models,'' \emph{arXiv preprint
  arXiv:2207.13038}, 2022.

\bibitem{md}
D.~Holz, ``Midjourney,'' \emph{https://www.midjourney.com/}, 2023.

\bibitem{attngan}
T.~Xu, P.~Zhang, Q.~Huang, H.~Zhang, Z.~Gan, X.~Huang, and X.~He, ``Attngan:
  Fine-grained text to image generation with attentional generative adversarial
  networks,'' in \emph{Proceedings of the IEEE conference on computer vision
  and pattern recognition}, 2018, pp. 1316--1324.

\bibitem{dalle2}
A.~Ramesh, P.~Dhariwal, A.~Nichol, C.~Chu, and M.~Chen, ``Hierarchical
  text-conditional image generation with clip latents,'' \emph{arXiv preprint
  arXiv:2204.06125}, vol.~1, no.~2, p.~3, 2022.

\bibitem{zhou2022towards}
Y.~Zhou, R.~Zhang, C.~Chen, C.~Li, C.~Tensmeyer, T.~Yu, J.~Gu, J.~Xu, and
  T.~Sun, ``Towards language-free training for text-to-image generation,'' in
  \emph{Proceedings of the IEEE/CVF Conference on Computer Vision and Pattern
  Recognition}, 2022, pp. 17\,907--17\,917.

\bibitem{bao2023one}
F.~Bao, S.~Nie, K.~Xue, C.~Li, S.~Pu, Y.~Wang, G.~Yue, Y.~Cao, H.~Su, and
  J.~Zhu, ``One transformer fits all distributions in multi-modal diffusion at
  scale,'' \emph{arXiv preprint arXiv:2303.06555}, 2023.

\bibitem{controlnet}
L.~Zhang, A.~Rao, and M.~Agrawala, ``Adding conditional control to
  text-to-image diffusion models,'' in \emph{Proceedings of the IEEE/CVF
  International Conference on Computer Vision}, 2023, pp. 3836--3847.

\bibitem{adam}
D.~P. Kingma and J.~Ba, ``Adam: A method for stochastic optimization,''
  \emph{arXiv preprint arXiv:1412.6980}, 2014.

\bibitem{fu2023scale}
J.~Fu, ``Scale guided hypernetwork for blind super-resolution image quality
  assessment,'' \emph{arXiv preprint arXiv:2306.02398}, 2023.

\bibitem{zhou2020blind}
W.~Zhou, Q.~Jiang, Y.~Wang, Z.~Chen, and W.~Li, ``Blind quality assessment for
  image superresolution using deep two-stream convolutional networks,''
  \emph{Information Sciences}, vol. 528, pp. 205--218, 2020.

\bibitem{zhou2019dual}
W.~Zhou, Z.~Chen, and W.~Li, ``Dual-stream interactive networks for
  no-reference stereoscopic image quality assessment,'' \emph{IEEE Transactions
  on Image Processing}, vol.~28, no.~8, pp. 3946--3958, 2019.

\bibitem{paszke2019pytorch}
A.~Paszke, S.~Gross, F.~Massa, A.~Lerer, J.~Bradbury, G.~Chanan, T.~Killeen,
  Z.~Lin, N.~Gimelshein, L.~Antiga \emph{et~al.}, ``Pytorch: An imperative
  style, high-performance deep learning library,'' \emph{Advances in neural
  information processing systems}, vol.~32, 2019.

\end{thebibliography}

\end{document}